\pdfoutput=1

\documentclass[11pt]{article}

\usepackage[title]{appendix}

\usepackage{acl}

\usepackage{times}
\usepackage{latexsym}

\usepackage[T1]{fontenc}

\usepackage[utf8]{inputenc}

\usepackage{microtype}

\usepackage{subcaption,graphicx,dblfloatfix}
\graphicspath{ {./img/} }

\usepackage{amsmath}
\usepackage{mathtools}
\usepackage{amssymb}
\usepackage{outline}

\usepackage{bm}      
\usepackage{booktabs} 
\usepackage{multirow} 

\newcommand{\figref}[1]{Figure~\ref{fig:#1}}
\newcommand{\tblref}[1]{Table~\ref{table:#1}}

\newcommand{\eqnref}[1]{Equation~\ref{eq:#1}}

\newcommand{\s}[1]{{\footnotesize $\pm #1$}}
\newcommand{\B}[1]{\textbf{#1}}
\newcommand{\U}[1]{\underline{#1}}

\definecolor{mypurple}{HTML}{9F00FF} 
\definecolor{myyellow}{HTML}{BA9500} 
\definecolor{myred}{HTML}{920500}
\definecolor{mygreen}{HTML}{3F7C0D}
\definecolor{myblue}{HTML}{3848D1}

\DeclareMathOperator*{\argmax}{arg\,max}

\title{KinyaBERT: a Morphology-aware Kinyarwanda Language Model}

\author{Antoine Nzeyimana \\
  University of Massachusetts Amherst \\
  \texttt{anthonzeyi@gmail.com} \\\And
  Andre Niyongabo Rubungo \\
  Polytechnic University of Catalonia \\
  \texttt{niyongabor.andre@gmail.com} \\}

\begin{document}
\maketitle
\setlength{\abovedisplayskip}{3pt}
\setlength{\belowdisplayskip}{3pt}





%
%
%
\begin{abstract}
Pre-trained language models such as BERT have been successful at tackling many natural language processing tasks. However, the unsupervised sub-word tokenization methods commonly used in these models (e.g., byte-pair encoding -- BPE) are sub-optimal at handling morphologically rich languages. Even given a morphological analyzer, naive sequencing of morphemes into a standard BERT architecture is inefficient at capturing morphological compositionality and expressing word-relative syntactic regularities. We address these challenges by proposing a simple yet effective two-tier BERT architecture that leverages a morphological analyzer and explicitly represents morphological compositionality. 
Despite the success of BERT, most of its evaluations have been conducted on high-resource languages, obscuring its applicability on low-resource languages. We evaluate our proposed method on the low-resource morphologically rich Kinyarwanda language, naming the proposed model architecture \textit{KinyaBERT}. A robust set of experimental results reveal that \textit{KinyaBERT} outperforms solid baselines by 2\% in F1 score on a named entity recognition task and by 4.3\% in average score of a machine-translated GLUE benchmark. \textit{KinyaBERT} fine-tuning has better convergence and achieves more robust results on multiple tasks even in the presence of translation noise.\footnote{\label{code1}Code and data are released at \url{https://github.com/anzeyimana/kinyabert-acl2022}}

\end{abstract}


\section{Introduction}

Recent advances in natural language processing (NLP) through deep learning have been largely enabled by vector representations (or embeddings) learned through language model pre-training~\cite{bengio2003neural, mikolov2013distributed, pennington2014glove, bojanowski2017enriching, peters2018deep, devlin2019bert}. Language models such as BERT~\citep{devlin2019bert} are pre-trained on large text corpora and then fine-tuned on downstream tasks, resulting in better performance on many NLP tasks. Despite attempts to make multilingual BERT models~\cite{conneau-etal-2020-unsupervised}, research has shown that models pre-trained on high quality monolingual corpora outperform multilingual models pre-trained on large Internet data~\cite{scheible2020gottbert, virtanen2019multilingual}. This has motivated many researchers to pre-train BERT models on individual languages rather than adopting the ``language-agnostic'' multilingual models. This work is partly motivated by the same findings, but also proposes an adaptation of the BERT architecture to address representational challenges that are specific to morphologically rich languages such as Kinyarwanda.

\begin{table*}[ht!]
\centering
\resizebox{\textwidth}{!}{%
\begin{tabular}{l l l l}
\toprule
\textbf{Word} & \textbf{Morphemes} & \textbf{Monolingual BPE} & \textbf{Multilingual BPE}  \rule{0pt}{2.8ex} \rule[-1.2ex]{0pt}{0pt}\\
\midrule
\textbf{twagezeyo} `\textit{we arrived there}' & \textbf{tu . a . \underline{ger} . ye . yo} & twag . ezeyo & \textbf{\_}twa . ge . ze . yo\\
\textbf{ndabyizeye} `\textit{I hope so}' & \textbf{n . ra . bi . \underline{izer} . ye} & ndaby . izeye & \textbf{\_} ndab . yiz . eye\\
\textbf{umwarimu} `\textit{teacher}' & \textbf{u . mu . \underline{arimu}} & umwarimu & \textbf{\_}um . wari . mu\\
\bottomrule
\end{tabular}
}
\caption{Comparison between morphemes and BPE-produced sub-word tokens. Stems are underlined.}
\label{table:bpe}
\vspace{-.1in}
\end{table*}

In order to handle rare words and reduce the vocabulary size, BERT-like models use statistical sub-word tokenization algorithms such as byte pair encoding (BPE)~\cite{sennrich2015neural}. While these techniques have been widely used in language modeling and machine translation, they are not optimal for morphologically rich languages~\citep{klein-tsarfaty-2020-getting}. In fact, sub-word tokenization methods that are solely based on surface forms, including BPE and character-based models, cannot capture all morphological details. This is due to morphological alternations \cite{muhirwe2007computational} and non-concatenative morphology~\citep{mccarthy1981prosodic} that are often exhibited by morphologically rich languages. For example, as shown in \tblref{bpe}, a BPE model trained on 390 million tokens of Kinyarwanda text cannot extract the true sub-word lexical units (i.e. morphemes) for the given words. This work addresses the above problem by proposing a language model architecture that explicitly represents most of the input words with morphological parses produced by a morphological analyzer. In this architecture BPE is only used to handle words which cannot be directly decomposed by the morphological analyzer such as misspellings, proper names and foreign language words.

Given the output of a morphological analyzer, a second challenge is in how to incorporate the produced morphemes into the model. One naive approach is to feed the produced morphemes to a standard transformer encoder as a single monolithic sequence. This approach is used by~\citet{mohseni-tebbifakhr-2019-morphobert}. One problem with this method is that mixing sub-word information and sentence-level tokens in a single sequence does not encourage the model to learn the actual morphological compositionality and express word-relative syntactic regularities. We address these issues by proposing a simple yet effective two-tier transformer encoder architecture. The first tier encodes morphological information, which is then transferred to the second tier to encode sentence level information. We call this new model architecture KinyaBERT because it uses BERT's masked language model objective for pre-training and is evaluated on the morphologically rich Kinyarwanda language.

This work also represents progress in low resource NLP. Advances in human language technology are most often evaluated on the main languages spoken by major economic powers such as English, Chinese and European languages. This has exacerbated the language technology divide between the highly resourced languages and the underrepresented languages. It also hinders progress in NLP research because new techniques are mostly evaluated on the mainstream languages and some NLP advances become less informed of the diversity of the linguistic phenomena~\citep{bender2019benderrule}. Specifically, this work provides the following research contributions:

\begin{itemize}
    \item A simple yet effective two-tier BERT architecture for representing morphologically rich languages.
    \item New evaluation datasets for Kinyarwanda language including a machine-translated subset of the GLUE benchmark~\cite{wang2018glue} and a news categorization dataset.
    \item Experimental results which set a benchmark for future studies on Kinyarwanda language understanding, and on using machine-translated versions of the GLUE benchmark.
    \item Code and datasets are made publicly available for reproducibility\textsuperscript{\ref{code1}}.
\end{itemize}

\section{Morphology-aware Language Model}

\begin{figure*}[!ht]
 \centering
    \includegraphics[scale=0.6]{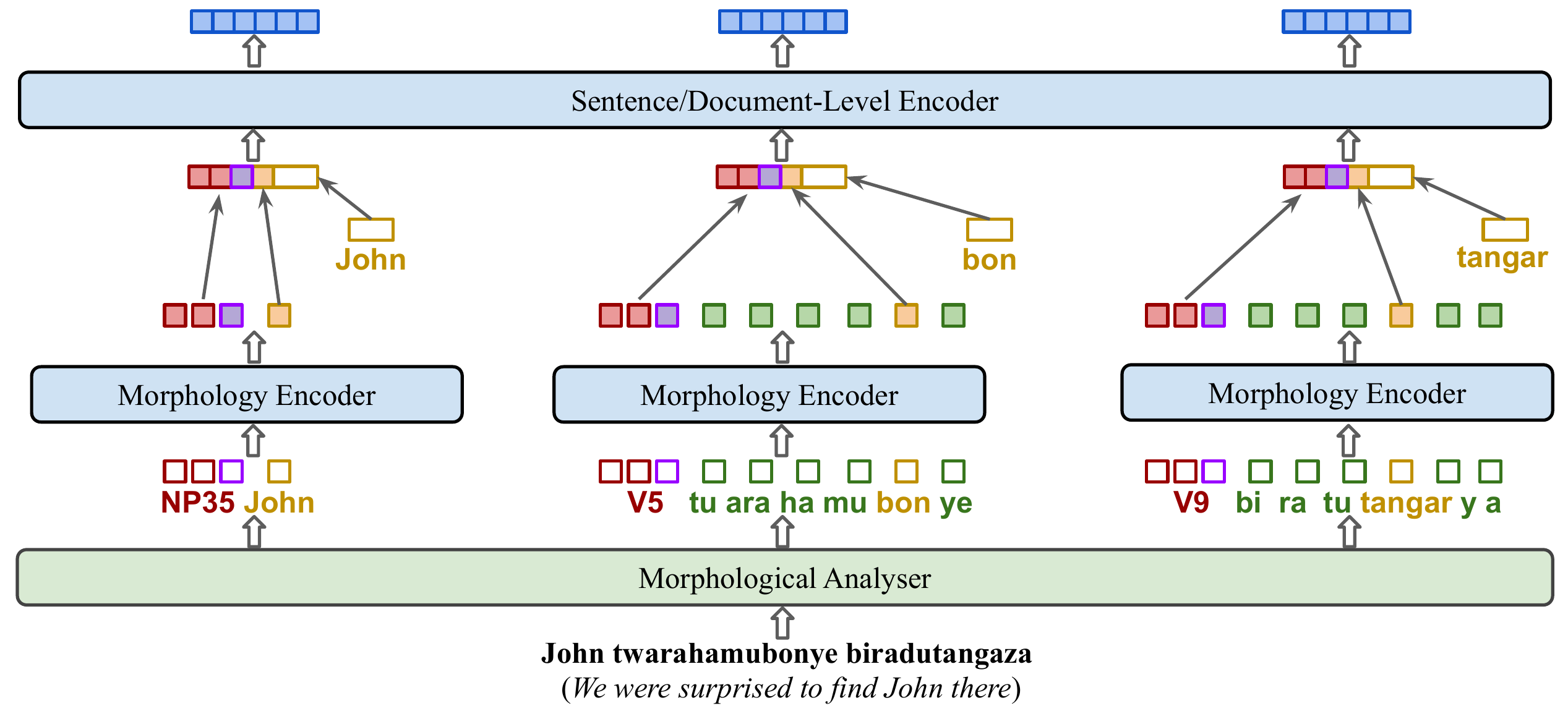}
 \captionof{figure}{\label{fig:kinyabert} KinyaBERT model architecture: Encoding of the sentence 'John twarahamusanze biradutangaza' (\textit{We were surprised to find John there}). The morphological analyzer produces morphemes for each word and assigns a POS tag to it. The two-tier transformer model then generates contextualized embeddings (\B{\textcolor{myblue}{blue}} vectors at the top). The \B{\textcolor{myred}{red}} colored embeddings correspond to the POS tags, \B{\textcolor{myyellow}{yellow}} is for the stem embeddings, \B{\textcolor{mygreen}{green}} is for the variable length affixes while the \B{\textcolor{mypurple}{purple}} embeddings correspond to the affix set.}
 \vspace{-.1in}
\end{figure*}

Our modeling objective is to be able to express morphological compositionality in a Transformer-based~\citep{vaswani2017attention} language model. For morphologically rich languages such as Kinyarwanda, a set of morphemes (typically a stem and a set of functional affixes) combine to produce a word with a given surface form. This requires an alternative to the ubiquitous BPE tokenization, through which exact sub-word lexical units (i.e. morphemes) are used. For this purpose, we use a morphological analyzer which takes a sentence as input and, for every word, produces a stem, zero or more affixes and assigns a part of speech (POS) tag to each word. This section describes how this morphological information is obtained and then integrated in a two-tier transformer architecture (\figref{kinyabert}) to learn morphology-aware input representations.

\subsection{Morphological Analysis and Part-of-Speech Tagging}

Kinyarwanda, the national language of Rwanda, is one of the major Bantu languages~\citep{nurse2006bantu} spoken in central and eastern Africa. Kinyarwanda has 16 noun classes. Modifiers (demonstratives, possessives, adjectives, numerals) carry a class marking morpheme that agrees with the main noun class. The verbal morphology~\citep{nzeyimana-2020-morphological} also includes subject and object markers that agree with the class of the subject or object. This agreement therefore enables users of the language to approximately disambiguate referred entities based on their classes. We leverage this syntactic agreement property in designing our unsupervised POS tagger.

Our morphological analyzer for Kinyarwanda was built following finite-state two-level morphology principles~\citep{koskenniemi1983two, beesley2000finite, beesley2003finite}. For every inflectable word type, we maintain a morphotactics model using a directed acyclic graph (DAG) that represents the regular sequencing of morphemes. We effectively model all inflectable word types in Kinyarwanda which include verbals, nouns, adjectives, possessive and demonstrative pronouns, numerals and quantifiers.
The morphological analyzer also includes many hand-crafted rules for handling morphographemics and other linguistic regularities of the Kinyarwanda language. The morphological analyzer was independently developed and calibrated by native speakers as a closed source solution before the current work on language modeling. Similar to~\citet{nzeyimana-2020-morphological}, we use a classifier trained on a stemming dataset to disambiguate between competing outputs of the morphological analyzer. Furthermore, we improve the disambiguation quality by leveraging a POS tagger at the phrase level so that the syntactic context can be taken into consideration.

We devise an unsupervised part of speech tagging algorithm which we explain here. Let ${x=(x_1,x_2,x_3,...x_n)}$ be a sequence of tokens (e.g. words) to be tagged with a corresponding sequence of tags ${y=(y_1,y_2,y_3,...y_n)}$. A sample of actual POS tags used for Kinyarwanda is given in \tblref{pos_tags_examples} the Appendix. Using Bayes' rule, the optimal tag sequence $y^{*}$ is given by the following equation:

\vspace{-.2in}

\begin{equation}\label{eq:bayes}
    \begin{split}
    y^* & = \argmax_{y} P(y|x) \\ 
     & = \argmax_{y} \frac{P(x|y)P(y)}{P(x)} \\
     & = \argmax_{y} P(x|y)P(y)
    \end{split}
\end{equation}

A standard hidden Markov model (HMM) can decompose the result of \eqnref{bayes} using first order Markov assumption and independence assumptions into $P(x|y)=\prod_{t=1}^n P(x_t|y_t)$ and $P(y)=\prod_{t=1}^n P(y_t|y_{t-1})$. The tag sequence $y^*$ can then be efficiently decoded using the Viterbi algorithm~\cite{forney1973viterbi}. A better decoding strategy is presented below.

Inspired by \citet{tsuruoka2005bidirectional}, we devise a greedy heuristic for decoding $y^*$ using the same first order Markov assumptions but with bidirectional decoding.

First, we estimate the local emission probabilities $P(x_t|y_t)$ using a factored model given in the following equation:
\begin{equation}\label{eq:factored}
\begin{split}
& P(x_t|y_t) \propto \tilde{P}(x_t|y_t)\\
& \tilde{P}(x_t|y_t) = \tilde{P}_m(x_t|y_t)\tilde{P}_p(x_t|y_t)\tilde{P}_a(x_t|y_t) \\
\end{split}
\end{equation}

In \eqnref{factored}, $\tilde{P}_m(x_t|y_t)$ corresponds to the probability/score returned by a morphological disambiguation classifier, representing the uncertainty of the morphology of $x_t$. $\tilde{P}_p(x_t|y_t)$ corresponds to a local precedence weight between competing POS tags. These precedence weights are manually crafted through qualitative evaluation (See \tblref{pos_tags_examples} in Appendix for examples). $\tilde{P}_a(x_t|y_t)$ quantifies the local neighborhood syntactic agreement between Bantu class markers. When there are two or more agreeing class markers in neighboring words, the tagger should be more confident of the agreeing parts of speech. A basic agreement score can be the number of agreeing class markers within a window of seven words around a given candidate $x_t$. We manually designed a more elaborate set of agreement rules and their weights for different contexts. Therefore, the actual agreement score $\tilde{P}_a(x_t|y_t)$ is a weighted sum of the matched agreement rules. Each of the unnormalized measures $\tilde{P}$ in \eqnref{factored} is mapped to the $[0,1]$ range using a sigmoid function $\sigma(z|z_{A},z_{B})$ given in \eqnref{sigma}, where $z$ is the score of the measure and $[z_A,z_B]$ is its estimated active range.
\begin{equation}\label{eq:sigma}
\begin{split}
& \sigma(z|z_{A},z_{B}) =  [1+exp(-8\frac{z-z_{A}}{z_{B}-z_{A}})]^{-8} \\
\end{split}
\end{equation}

After estimating the local emission model, we greedily decode $y_t^* = \argmax{y_t} \tilde{P}(y_t|x)$ in decreasing order of $\tilde{P}(x_t|y_t)$ using a first order bidirectional inference of $\tilde{P}(y_t|x)$ as given in the following equation:

\begin{equation}\label{eq:infer}
\begin{split}
    & \tilde{P}(y_t|x) = \\
    & \begin{cases}
        \tilde{P}(x_t|y_t)\tilde{P}(y_t|y^*_{t-1},y^*_{t+1})\tilde{P}(y^*_{t-1}|x)\tilde{P}(y^*_{t+1}|x)\\ 
        \quad \quad \text{ if both } y^*_{t-1} \text{ and } y^*_{t+1} \text{ have been decoded;} \\
        \tilde{P}(x_t|y_t)\tilde{P}(y_t|y^*_{t-1})\tilde{P}(y^*_{t-1}|x)  \\ 
        \quad \quad \text{ if only } y^*_{t-1} \text{ has been decoded;} \\
        \tilde{P}(x_t|y_t)\tilde{P}(y_t|y^*_{t+1})\tilde{P}(y^*_{t+1}|x)   \\ 
        \quad \quad \text{ if only } y^*_{t+1} \text{ has been decoded;} \\
        \tilde{P}(x_t|y_t)\text{ otherwise}
    \end{cases}
\end{split}
\end{equation}

The first order transition measures $\tilde{P}(y_t|y_{t-1})$, $\tilde{P}(y_t|y_{t+1})$ and $\tilde{P}(y_t|y_{t-1},y_{t+1})$ are estimated using count tables computed over the entire corpus by aggregating local emission marginals ${\tilde{P}(y_t)=\sum_{x_t}\tilde{P}(x_t,y_t)}$ obtained through morphological analysis and disambiguation.

\subsection{Morphology Encoding}

The overall architecture of our model is depicted in \figref{kinyabert}. This is a two-tier transformer encoder architecture made of a token-level morphology encoder that feeds into a sentence/document-level encoder. The morphology encoder is made of a small transformer encoder that is applied to each analyzed token separately in order to extract its morphological features. The extracted morphological features are then concatenated with the token's stem embedding to form the input vector fed to the sentence/document encoder. The sentence/document encoder is made of a standard transformer encoder as used in other BERT models. The sentence/document encoder uses untied position encoding with relative bias as proposed in~\citet{ke2020rethinking}.

The input to the morphology encoder is a set of embedding vectors, three vectors relating to the part of speech, one for the stem and one for each affix when available. The transformer encoder operation is applied to these embedding vectors without any positional information. This is because positional information at the morphology level is inherent since no morpheme repeats and each morpheme always occupies a known (i.e. fixed) slot in the morphotactics model. The extracted morphological features are four encoder output vectors corresponding to the three POS embeddings and one stem embedding. Vectors corresponding to the affixes are left out since they are of variable length and the role of the affixes in this case is to be attended to by the stem and the POS tag so that morphological information can be captured. The four morphological output feature vectors are further concatenated with another stem embedding at the sentence level to form the input vector for the main sentence/document encoder.

The choice of this transformer-based architecture for morphology encoding  is motivated by two factors. First, \citet{zaheer2020big} has demonstrated the importance of having ``global tokens'' such as \texttt{[CLS]} token in BERT models. These are tokens that attend to all other tokens in the modeled sequence. These ``global tokens'' effectively encapsulate some ``meaning'' of the encoded sequence. Second, the POS tag and stem represent the high level information content of a word. Therefore, having the POS tag and stem embeddings be transformed into morphological features is a viable option. The POS tag and stem embeddings thus serve as the ``global tokens'' at the morphology encoder level since they attend to all other morphemes that can be associated with them.

In order to capture subtle morphological information, we make one of the three POS embeddings span an affix set vocabulary that is a subset of the all affixes power set. We form an affix set vocabulary $\mathcal{V}_a$ that is made of the $N$ most frequent affix combinations in the corpus. In fact, the morphological model of the language enforces constraints on which affixes can go together for any given part of speech, resulting in an affix set vocabulary that is much smaller than the power set of all affixes. Even with limiting the affix set vocabulary $\mathcal{V}_a$ to a fixed size, we can still map any affix combination to $\mathcal{V}_a$ by dropping zero or very few affixes from the combination. Note that the affix set embedding still has to attend to all morphemes at the morphology encoder level, making it adapt to the whole morphological context. The affix set embedding is depicted by the \B{\textcolor{mypurple}{purple}} units in \figref{kinyabert} and a sample of $\mathcal{V}_a$ is given in~\tblref{affix_set_examples} in the Appendix.

\subsection{Pre-training Objective}

Similar to other BERT models, we use a masked language model objective. Specifically, 15\% of all tokens in the training set are considered for prediction, of which 80\% are replaced with \texttt{[MASK]} tokens, 10\% are replaced with random tokens and 10\% are left unchanged. When prediction tokens are replaced with \texttt{[MASK]} or random tokens, the corresponding affixes are randomly omitted 70\% of the time or left in place for 30\% of the time, while the units corresponding to POS tags and affix sets are also masked. The pre-training objective is then to predict stems and the associated affixes for all tokens considered for prediction using a two-layer feed-forward module on top of the encoder output. 

For the affix prediction task, we face a multi-label classification problem where for each prediction token, we predict a variable number of affixes. In our experiments, we tried two methods. For one, we use the Kullback–Leibler (KL) divergence\footnote{\url{https://en.wikipedia.org/wiki/Kullback\%E2\%80\%93Leibler_divergence}} loss function to solve a regression task of predicting the $N$-length affix distribution vector. For this case, we use a target affix probability vector $\bm{a_t} \in \mathbb{R}^N$ in which each target affix index is assigned $\frac{1}{m}$ probability and $0$ probability for non-target affixes. Here $m$ is the number of affixes in the word to be predicted and $N$ is the total number of all affixes. We call this method ``Affix Distribution Regression'' (ADR) and model variant KinyaBERT$_{ADR}$. Alternatively, we use cross entropy loss and just predict the affix set associated with the prediction word; we call this method ``Affix Set Classification'' (ASC) and the model variant KinyaBERT$_{ASC}$.


\section{Experiments}

In order to evaluate the proposed architecture, we pre-train KinyaBERT (101M parameters for KinyaBERT$_{ADR}$ and 105M for KinyaBERT$_{ASC}$) on a 2.4 GB of Kinyarwanda text along with 3 baseline BERT models. The first baseline is a BERT model pre-trained on the same Kinyarwanda corpus and with the same position encoding~\cite{ke2020rethinking}, same batch size and pre-training steps, but using the standard BPE tokenization. We call this first baseline model BERT$_{BPE}$ (120M parameters). The second baseline is a similar BERT model pre-trained on the same Kinyarwanda corpus but tokenized by a morphological analyzer. For this model, the input is just a sequence of morphemes, in a similar fashion to~\citet{mohseni-tebbifakhr-2019-morphobert}. We call this second baseline model BERT$_{MORPHO}$ (127M parameters). For BERT$_{MORPHO}$, we found that predicting 30\% of the tokens achieves better results than using 15\% because of the many affixes generated. The third baseline is XLM-R~\cite{conneau-etal-2020-unsupervised} (270M parameters) which is pre-trained on 2.5 TB of multilingual text. We evaluate the above models by comparing their performance on downstream NLP tasks.

\begin{table}[ht!]
\centering
\begin{tabular}{l r}
\toprule
Language & Kinyarwanda \\
Publication Period & 2011 - 2021 \\
Websites/Sources & 370 \\
Documents/Articles & 840K \\
Sentences & 16M \\
Tokens/Words & 390M \\
Text size & 2.4 GB \\
\bottomrule
\end{tabular}
\caption{Summary of the pre-training corpus.}
\label{table:pretrain_corpus_stat}
 \vspace{-.2in}
\end{table}

\subsection{Pre-training details}

KinyaBERT model was implemented using Pytorch version 1.9. The morphological analyzer and POS tagger were implemented in a shared library using POSIX C. Morphological parsing of the corpus was performed as a pre-processing step, taking 20 hours to segment the 390M-token corpus on an 12-core desktop machine. Pre-training was performed using RTX 3090 and RTX 2080Ti desktop GPUs. Each KinyaBERT model takes on average 22 hours to train for 1000 steps on one RTX 3090 GPU or 29 hours on one RTX 2080Ti GPU. Baseline models (BERT$_{BPE}$ and BERT$_{MORPHO}$) were pre-trained on cloud tensor processing units (TPU v3-8 devices each with 128 GB memory) using PyTorch/XLA\footnote{\url{https://github.com/pytorch/xla/}} package and a TPU-optimized fairseq toolkit~\cite{ott-etal-2019-fairseq}. Pre-training on TPU took 2.3 hours per 1000 steps. The baselines were trained on TPU because there were no major changes needed to the existing RoBERTA (base) architecture implemented in fairseq and the TPU resources were available and efficient. In all cases, pre-training batch size was set to 2560 sequences, with maximum 512 tokens in each sequence. The maximum learning rates was set to $4\times10^{-4}$ which is achieved after 2000 steps and then linearly  decays. Our main results and ablation results were obtained from models pre-trained for 32K steps in all cases. Other pre-training details, model architectural dimensions and other hyper-parameters are given in the Appendix.




\begin{table*}[ht!]
\centering
\resizebox{\textwidth}{!}{%
\begin{tabular}{l c c c c c c c c c c c c c}
\toprule
\B{Task:}              & \B{MRPC}    & \B{QNLI}    & \B{RTE}    & \B{SST-2}  & \B{STS-B}      & \B{WNLI} \\
\B{\#Train examples:}  & \B{3.4K}    & \B{104.7K}  & \B{2.5K}   & \B{67.4K}  & \B{5.8K}       & \B{0.6K} \\
\B{Translation score:} & \B{2.7/4.0}  & \B{2.9/4.0}  & \B{3.0/4.0} & \B{2.7/4.0} & \B{3.1/4.0}     & \B{2.9/4.0} \\
\midrule
\B{Model} & \multicolumn{6}{c}{\B{Validation Set}}\\
\midrule
XLM-R          & 84.2/78.3\s{0.8/1.0}     & 79.0\s{0.3}     & 58.4\s{3.2}     & 78.7\s{0.6}     & 77.7/77.8\s{0.7/0.6}     & 55.4\s{2.0} \\
BERT$_{BPE}$       & 83.3/76.6\s{0.8/1.4}     & 81.9\s{0.2}     & 59.2\s{1.5}     & 80.1\s{0.4}     & 75.6/75.7\s{7.8/7.3}     & 55.4\s{1.9} \\
BERT$_{MORPHO}$    & 84.3/77.4\s{0.6/1.1}     & 81.6\s{0.2}     & 59.2\s{1.5}     & 81.6\s{0.5}     & 76.8/77.0\s{0.8/0.7}     & 54.2\s{2.5} \\
KinyaBERT$_{ADR}$ & \B{87.1/82.1}\s{0.5/0.7} & 81.6\s{0.1}     & 61.8\s{1.4}     & 81.8\s{0.6}     & 79.6/79.5\s{0.4/0.3}     & 54.5\s{2.2} \\
KinyaBERT$_{ASC}$ & 86.6/81.3\s{0.5/0.7}     & \B{82.3}\s{0.3} & \B{64.3}\s{1.4} & \B{82.4}\s{0.5} & \B{80.0/79.9}\s{0.5/0.5} & \B{56.2}\s{0.8} \\
\midrule
\textbf{Model} & \multicolumn{6}{c}{\textbf{Test Set}}\\
\midrule
XLM-R          & 82.6/76.0\s{0.6/0.6}     & 78.1\s{0.3}     & 56.4\s{3.2}     & 76.3\s{0.4}     & 69.5/68.9\s{1.0/1.1}     & 63.7\s{3.9} \\
BERT$_{BPE}$       & 82.8/76.2\s{0.6/0.8}     & 81.1\s{0.3}     & 55.6\s{2.8}     & 79.1\s{0.4}     & 68.9/67.8\s{1.8/1.7}     & 63.4\s{4.1} \\
BERT$_{MORPHO}$    & 82.7/75.4\s{0.8/1.3}     & 80.8\s{0.4}     & 56.7\s{1.0}     & 80.7\s{0.5}     & 68.9/67.8\s{1.5/1.3}     & \U{65.0}\s{0.3} \\
KinyaBERT$_{ADR}$ & 84.4/\B{78.7}\s{0.5/0.6} & 81.2\s{0.3}     & 58.1\s{1.1}     & 80.9\s{0.5}     & 73.2/72.0\s{0.4/0.3}     & \U{65.1}\s{0.0} \\
KinyaBERT$_{ASC}$ & \B{84.6}/78.4\s{0.2/0.3} & \B{82.2}\s{0.6} & \B{58.8}\s{0.7} & \B{81.4}\s{0.6} & \B{74.5/73.5}\s{0.2/0.2} & \U{65.0}\s{0.2} \\
\bottomrule
\end{tabular}
}
\caption{Performance results on the machine translated GLUE benchmark~\cite{wang2018glue}. The translation score is the sample average translation quality score assigned by volunteers. For MRPC, we report accuracy and F1. For STS-B, we report Pearson and Spearman correlations. For all others, we report accuracy. The best results are shown in \B{bold} while equal top results are \U{underlined}.}
\label{table:glue_results}
 \vspace{-.1in}
\end{table*}


\begin{table}[ht!]
\centering
\resizebox{\columnwidth}{!}{%
\begin{tabular}{l c c}
\toprule
\B{Task:} & \multicolumn{2}{c}{\B{NER}} \\
\B{\#Train examples:} & \multicolumn{2}{c}{\B{2.1K}} \\
\midrule
\B{Model} & \B{Validation Set} & \B{Test Set}\\
\midrule
XLM-R          & 80.3\s{1.0}     & 71.8\s{1.5} \\
BERT$_{BPE}$       & 83.4\s{0.9}     & 74.8\s{0.8} \\
BERT$_{MORPHO}$    & 83.2\s{0.9}     & 72.8\s{0.9} \\
KinyaBERT$_{ADR}$ & \B{87.1}\s{0.8} & \B{77.2}\s{1.0} \\
KinyaBERT$_{ASC}$ & 86.2\s{0.4}     & 76.3\s{0.5} \\
\bottomrule
\end{tabular}
}
\caption{Micro average F1 scores on Kinyarwanda NER task~\cite{adelani-etal-2021-masakhaner}.}
\label{table:ner_results}
\end{table}


\begin{table}[ht!]
\centering
\resizebox{\columnwidth}{!}{%
\begin{tabular}{l c c}
\toprule
\B{Task:} & \multicolumn{2}{c}{\B{NEWS}} \\
\B{\#Train examples:} & \multicolumn{2}{c}{\B{18.0K}} \\
\midrule
\B{Model} & \B{Validation Set} & \B{Test Set}\\
\midrule
XLM-R          & 83.8\s{0.3}     & 84.0\s{0.2} \\
BERT$_{BPE}$       & 87.6\s{0.4}     & \B{88.3}\s{0.3} \\
BERT$_{MORPHO}$    & 86.9\s{0.4}     & 86.9\s{0.3} \\
KinyaBERT$_{ADR}$ & \B{88.8}\s{0.3} & 88.0\s{0.3} \\
KinyaBERT$_{ASC}$ & 88.4\s{0.3}     & 88.0\s{0.2} \\
\bottomrule
\end{tabular}
}
\caption{Accuracy results on Kinyarwanda NEWS categorization task.}
\label{table:news_results}
 \vspace{-.1in}
\end{table}

\subsection{Evaluation tasks}

\B{Machine translated GLUE benchmark} -- The General Language Understanding Evaluation (GLUE) benchmark~\cite{wang2018glue} has been widely used to evaluate pre-trained language models. In order to assess KinyaBERT performance on such high level language tasks, we used Google Translate API to translate a subset of the GLUE benchmark (MRPC, QNLI, RTE, SST-2, STS-B and WNLI tasks) into Kinyarwanda. CoLA task was left because it is English-specific. MNLI and QQP tasks were also not translated because they were too expensive to translate with Google's commercial API. While machine translation adds more noise to the data, evaluating on this dataset is still relevant because all models compared have to cope with the same noise. To understand this translation noise, we also run user evaluation experiments, whereby four volunteers proficient in both English and Kinyarwanda evaluated a random sample of 6000 translated GLUE examples, and assigned a score to each example on a scale from 1 to 4 (See \tblref{glue_scores} in Appendix). These scores help us characterize the noise in the data and contextualize our results with regards to other GLUE evaluations. Results on these GLUE tasks are shown in \tblref{glue_results}.

\B{Named entity recognition (NER)} -- We use the Kinyarwanda subset of the MasakhaNER dataset~\cite{adelani-etal-2021-masakhaner} for NER task. This is a high quality NER dataset annotated by native speakers for major African languages including Kinyarwanda. The task requires predicting four entity types: Persons (PER), Locations (LOC), Organizations (ORG), and date and time (DATE). Results on this NER task are presented in \tblref{ner_results}.

\B{News Categorization Task (NEWS)} -- For a document classification experiment, we collected a set of categorized news articles from seven major news websites that regularly publish in Kinyarwanda. The articles were already categorized, so no more manual labeling was needed. This dataset is similar to~\citet{niyongabo2020kinnews}, but in our case, we limited the number collected articles per category to 3000 in order to have a more balanced label distribution (See \tblref{rw_news} in the Appendix). The final dataset contains a total of 25.7K articles spanning 12 categories and has been split into training, validation and test sets in the ratios of 70\%, 5\% and 25\% respectively. Results on this NEWS task are presented in \tblref{news_results}.

For each evaluation task, we use a two-layer feed-forward network on top of the sentence encoder as it is typically done in other BERT models. The fine-tuning hyper-parameters are presented in~\tblref{finetune_hyperpar} in the Appendix.

\subsection{Main results}

The main results are presented in \tblref{glue_results}, \tblref{ner_results}, and \tblref{news_results}. Each result is the average of 10 independent fine-tuning runs. Each average result is shown with the standard deviation of the 10 runs. Except for XLM-R, all other models are pre-trained on the same corpus (See \tblref{pretrain_corpus_stat}) for 32K steps using the same hyper-parameters.

On the GLUE task, KinyaBERT$_{ASC}$ achieves 4.3\% better average score than the strongest baseline. KinyaBERT$_{ASC}$ also leads to more robust results on multiple tasks. It is also shown that having just a morphological analyzer is not enough: BERT$_{MORPHO}$ still under-performs even though it uses morphological tokenization. Multi-lingual XLM-R achieves least performance in most cases, possibly because it was not pre-trained on Kinyarwanda text and uses inadequate tokenization.

On the NER task, KinyaBERT$_{ADR}$ achieves best performance, about 3.2\% better average F1 score than the strongest baseline. One of the architectural differences between KinyaBERT$_{ADR}$ and KinyaBERT$_{ASC}$ is that KinyaBERT$_{ADR}$ uses three POS tag embeddings while KinyaBERT$_{ASC}$ uses two. Assuming that POS tagging facilitates named entity recognition, this empirical result suggests that increasing the amount of POS tag information in the model, possibly through diversification (i.e. multiple POS tag embedding vectors per word), can lead to better NER performance.

The NEWS categorization task resulted in differing performances between validation and test sets. This may be a result that solving such task does not require high level language modeling but rather depends on spotting few keywords. Previous research on a similar task~\cite{niyongabo2020kinnews} has shown that simple classifiers based on TF-IDF features suffice to achieve best performance.

The morphological analyzer and POS tagger inherently have some level of noise because they do not always perform with perfect accuracy. While we did not have a simple way of assessing the impact of this noise in this work, we can logically expect that the lower the noise the better the results could be. Improving the morphological analyzer and POS tagger and quantitatively evaluating its accuracy is part of future work. Even though our POS tagger uses heuristic methods and was evaluated mainly through qualitative exploration, we can still see its positive impact on the pre-trained language model. We did not use previous work on Kinyarwanda POS tagging because it is largely different from this work in terms of scale, tag dictionary and dataset size and availability.

\begin{figure*}[ht!]
 \centering
\resizebox{\textwidth}{!}{%
    \includegraphics{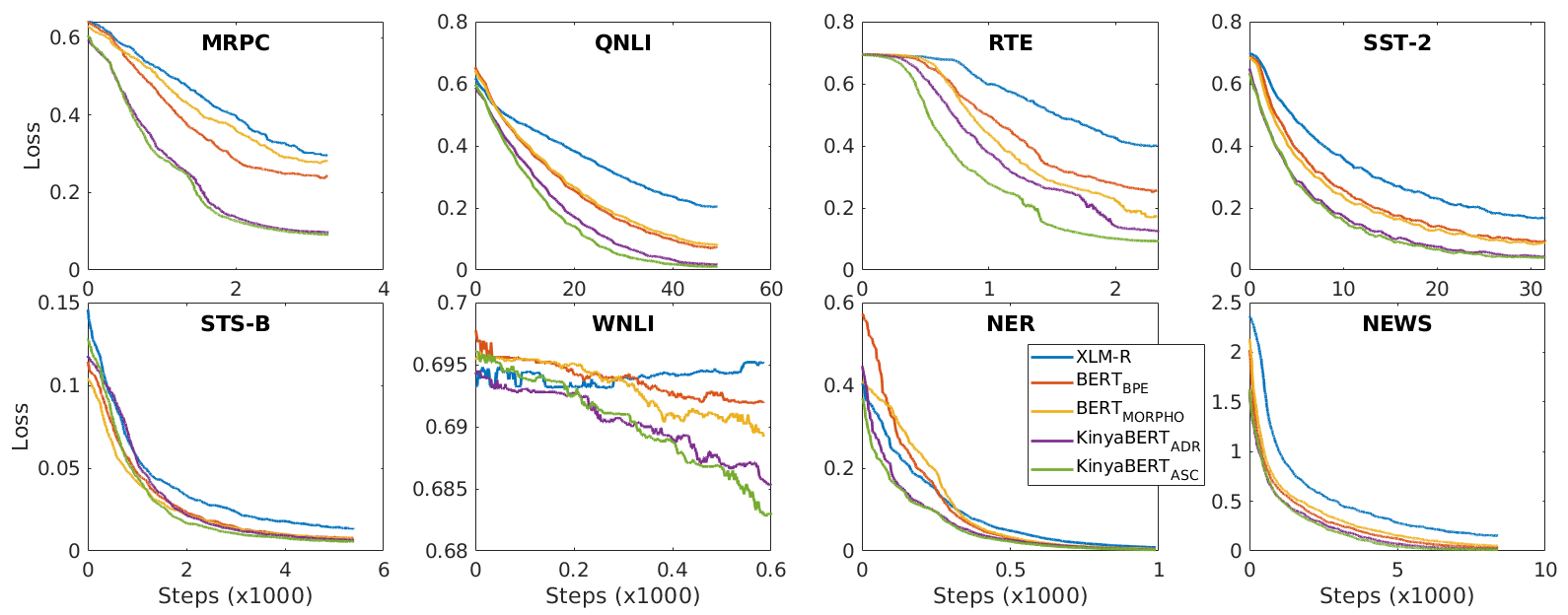}
    }
 \captionof{figure}{Comparison of fine-tuning loss curves between KinyaBERT and baselines on the evaluation tasks. KinyaBERT$_{ASC}$ achieves the best convergence in most cases, indicating better effectiveness of its model architecture and pre-training objective.}
 \label{fig:finetune}
  \vspace{-.1in}
\end{figure*}

We plot the learning curves during fine-tuning process of KinyaBERT and the baselines. The results in \figref{finetune} indicate that KinyaBERT fine-tuning has better convergence across all tasks. Additional results also show that positional attention~\cite{ke2020rethinking} learned by KinyaBERT has more uniform and smoother relative bias while BERT$_{BPE}$ and BERT$_{MORPHO}$ have more noisy relative positional bias (See \figref{pos_bias} in Appendix). This is possibly an indication that KinyaBERT allows learning better word-relative syntactic regularities. However, this aspect needs to be investigated more systematically in future research.

While the main sentence/document encoder of KinyaBERT is equivalent to a standard BERT ``BASE'' configuration on top of a small morphology encoder, overall, the model actually decreases the number of parameters by more than 12\% through embedding layer savings. This is because using morphological representation reduces the vocabulary size. Using smaller embedding vectors at the morphology encoder level also significantly reduces the overall number of parameters. \tblref{vocab_sizes} in Appendix shows the vocabulary sizes and parameter count of KinyaBERT in comparison to the baselines. While the sizing of the embeddings was done essentially to match BERT ``BASE'' configuration, future studies can shed more light on how different model sizes affect performance.

\subsection{Ablation study}


\begin{table*}[ht!]
\centering
\resizebox{\textwidth}{!}{%
\begin{tabular}{l c c c c c c c c c c c c c c c}
\toprule
\B{Task:}                & \B{MRPC}   & \B{QNLI}   & \B{RTE}   & \B{SST-2} & \B{STS-B}     & \B{WNLI} & \B{NER} & \B{NEWS} \\
\midrule
\B{Morphology$\rightarrow$Prediction} & \multicolumn{6}{c}{\B{Validation Set}}\\
\midrule
AFS$\rightarrow$STEM+ASC      & 86.6/81.3     & \B{82.3} & \B{64.3} & \B{82.4} & \B{80.0/79.9} & \U{56.2} & 86.2     & 88.4 \\
POS$\rightarrow$STEM+ADR      & \B{87.1/82.1} & 81.6     & 61.8     & 81.8     & 79.6/79.5     & 54.5     & \B{87.1} & \B{88.8} \\
AVG$\rightarrow$STEM+ADR      & 85.5/80.3     & 81.4     & 63.0     & 82.1     & 79.6/79.5     & \U{55.8} & 86.6     & 88.3 \\
STEM$\rightarrow$STEM & 86.4/81.5     & 80.4     & 63.4     & 77.5     & 79.7/79.5     & 50.4     & 86.6     & 88.0 \\
\midrule
\textbf{Morphology$\rightarrow$Prediction} & \multicolumn{6}{c}{\textbf{Test Set}}\\
\midrule
AFS$\rightarrow$STEM+ASC      & \B{84.6}/78.4 & \B{82.2} & 58.8     & \B{81.4} & \B{74.5/73.5} & \U{65.0} & 76.3     & 88.0 \\
POS$\rightarrow$STEM+ADR      & 84.4/\B{78.7} & 81.2     & 58.1     & 80.9     & 73.2/72.0     & \U{65.1} & \B{77.2} & 88.0 \\
AVG$\rightarrow$STEM+ADR      & 84.0/78.2     & 81.7     & \U{59.4} & 80.7     & 73.6/72.6     & \U{65.0} & 76.9     & 88.2 \\
STEM$\rightarrow$STEM & 84.2/78.6     & 80.3     & \U{59.8} & 77.5     & 73.3/72.0     & 59.6     & 76.4     & \B{88.4} \\
\bottomrule
\end{tabular}
}
\caption{Ablation results: each result is an average of 10 independent fine-tuning runs. Metrics, dataset sizes and noise statistics are the same as for the main results in \tblref{glue_results}, \tblref{ner_results} and \tblref{news_results}.}
\label{table:results_ablation}
\end{table*}

We conducted an ablation study to clarify some of the design choices made for KinyaBERT architecture. We make variations along two axes: (i) morphology input and (ii) pre-training task, which gave us four variants that we pre-trained for 32K steps and evaluated on the same downstream tasks.

\begin{itemize}
\item \B{AFS$\rightarrow$STEM+ASC}: Morphological features are captured by two POS tag and one affix set vectors. We predict both the stem and affix set. This corresponds to KinyaBERT$_{ASC}$ presented in the main results.
\item \B{POS$\rightarrow$STEM+ADR}: Morphological features are carried by three POS tag vectors and we predict the stem and affix probability vector. This corresponds to KinyaBERT$_{ADR}$.
\item \B{AVG$\rightarrow$STEM+ADR}: Morphological features are captured by two POS tag vectors and the pointwise average of affix hidden vectors from the morphology encoder. We predict the stem and affix probability vector.
\item \B{STEM$\rightarrow$STEM}: We omit the morphology encoder and train a model with only the stem parts without affixes and only predict the stem.
\end{itemize}

Ablation results presented in \tblref{results_ablation} indicate that using affix sets for both morphology encoding and prediction gives better results for many GLUE tasks. The under-performance of ``STEM$\rightarrow$STEM'' on high resource tasks (QNLI and SST-2) is an indication that morphological information from affixes is important. However, the utility of this information depends on the task as we see mixed results on other tasks.

Due to a large design space for a morphology-aware language model, there are still a number of other design choices that can be explored in future studies. One may vary the amount of POS tag embeddings used, vary the size affix set vocabulary or the dimension of the morphology encoder embeddings. One may also investigate the potential of other architectures for the morphology encoder, such as convolutional networks. Our early attempt of using recurrent neural networks (RNNs) for the morphology encoder was abandoned because it was too slow to train.


\section{Related Work}

BERT-variant pre-trained language models (PLMs) were initially pre-trained on monolingual high-resource languages. Multilingual PLMs that include both high-resource and low-resource languages have also been introduced~\cite{devlin2019bert, conneau-etal-2020-unsupervised, xue2020mt5,chung2020rethinking}. However, it has been found that these multilingual models are biased towards high-resource languages and use fewer low quality and uncleaned low-resource data~\cite{kreutzer2022quality}. The included low-resource languages are also very limited because they are mainly sourced from Wikipedia articles, where languages with few articles like Kinyarwanda are often left behind~\cite{joshi2020state, nekoto2020participatory}.

\citet{joshi2020state} classify the state of NLP for Kinyarwanda as ``Scraping-By'', meaning it has been mostly excluded from previous NLP research, and require the creation of dedicated resources and models. Kinyarwanda has been studied mostly in descriptive linguistics~\cite{kimenyi1976subjectivization,kimenyi1978aspects,kimenyi1978relational,kimenyi1988passiveness,jerrolocative}. Few recent NLP works on Kinyarwanda include Morphological Analysis~\cite{muhirwe2009morphological,nzeyimana-2020-morphological}, Text Classification~\cite{niyongabo2020kinnews}, Named Entity Recognition~\cite{rijhwani2020soft,adelani-etal-2021-masakhaner,saleva2021mining}, POS tagging~\cite{garrette2013learning,garrette2013real,duong2014can,fang2016learning,cardenas2019grounded}, and Parsing~\cite{sun2014parsing,mielens2015parse}. There is no prior study on pre-trained language modeling for Kinyarwanda.

There are very few works on monolingual PLMs for African languages. To the best of our knowledge there is currently only AfriBERT~\cite{ralethe2020adaptation} that has been pre-trained on Afrikaans, a language spoken in South Africa. In this paper, we aim to increase the inclusion of African languages in NLP community by introducing a PLM for Kinyarwanda. Differently to the previous works~(see \tblref{monobert2} in Appendix) which solely pre-trained unmodified BERT models, we propose an improved BERT architecture for morphologically rich languages.

Recently, there has been a research push to improve sub-word tokenization by adopting character-based models~\citep{ma2020charbert, clark2021canine}. While these methods are promising for the “language-agnostic” case, they are still solely based on the surface form of words, and thus have the same limitations as BPE when processing morphologically rich languages. We leave it to future research to empirically explore how these character-based methods compare to morphology-aware models.

\color{black}

\section{Conclusion}

This work demonstrates the effectiveness of explicitly incorporating morphological information in language model pre-training. The proposed two-tier Transformer architecture allows the model to represent morphological compositionality. Experiments conducted on Kinyarwanda, a low resource morphologically rich language, reveal significant performance improvement on several downstream NLP tasks when using the proposed architecture. These findings should motivate more research into morphology-aware language models.

\section*{Acknowledgements}

This work was supported with Cloud TPUs from Google's TPU Research Cloud (TRC) program and Google Cloud Research Credits  with the award GCP19980904. We also thank the anonymous reviewers for their insightful feedback.

\bibliography{acl}
\bibliographystyle{acl_natbib}

\clearpage

\begin{appendices}

\section{Data Tables, Hyper-parameters \& Additional results}

\begin{table}[h!]
\centering
\begin{tabular}{l r}
\toprule
\textbf{Module} & \textbf{Values}\\
\midrule
\textbf{Morphology Encoder:} & \\
\midrule
Number of Layers & 4\\
Attention heads &  4\\
Hidden Size & 128\\
Attention head size &  32\\
FFN inner hidden size &  512\\
Morphological embedding size & 128 \\
\midrule
\textbf{Sentence/Document Encoder:} & \\
\midrule
Number of Layers & 12\\
Attention heads &  12\\
Hidden Size & 768\\
Attention head size &  64\\
FFN inner hidden size &  3072\\
Stem embedding size & 256 \\
\bottomrule
\end{tabular}
\caption{KinyaBERT Architectural dimensions.}
\label{table:arch}
\end{table}

\begin{table}[ht!]
\centering
\begin{tabular}{l r}
\toprule
\textbf{Model} (\textbf{\#Params}) & \textbf{Vocab. Size}\\
\midrule
\textbf{XLM-R} (\textbf{270M}): & \\
Sentence-Piece tokens & 250K\\
\midrule
\textbf{BERT$_{BPE}$} (\textbf{120M}): & \\
BPE Tokens & 43K\\
\midrule
\textbf{BERT$_{MORPHO}$} (\textbf{127M}): & \\
Morphemes \& BPE Tokens & 51K\\
\midrule
\textbf{KinyaBERT$_{ADR}$} (\textbf{101M}): & \\
Stems \& BPE Tokens & 34K\\
Affixes &  0.3K\\
POS Tags & 0.2K\\
\midrule
\textbf{KinyaBERT$_{ASC}$} (\textbf{105M}): & \\
Stems \& BPE Tokens & 34K\\
Affix sets  & 34K\\
Affixes &  0.3K \\
POS Tags & 0.2K \\
\bottomrule
\end{tabular}
\caption{Vocabulary sizes for embedding layers.}
\label{table:vocab_sizes}
\end{table}

\newpage

\begin{table}[ht!]
\centering
\begin{tabular}{l r}
\toprule
\textbf{Hyper-parameter} & \textbf{Values}\\
\midrule
\midrule
Dropout &  0.1\\
Attention Dropout & 0.1\\
Warmup Steps & 2K\\
Max Steps &  200K\\
Weight Decay &  0.01\\
Learning Rate Decay &  Linear\\
Peak Learning Rate & 4e-4\\
Batch Size & 2560\\
Optimizer & LAMB\\
Adam $\epsilon$ & 1e-6 \\
Adam $\beta_{1}$ & 0.90\\
Adam $\beta_{2}$ & 0.98\\
Gradient Clipping & 0\\
\bottomrule
\end{tabular}
\caption{Pre-training hyper-parameters}
\label{table:pretrain_hyperpar}
\end{table}

\begin{table}[h!]
\centering
\begin{tabular}{l r}
\toprule
\textbf{Category} & \textbf{\#Articles}\\
\midrule
entertainment & 3000\\
sports & 3000\\
security & 3000\\
economy & 3000\\
health & 3000\\
politics & 3000\\
religion & 2020\\
development & 1813\\
technology & 1105\\
culture & 994\\
relationships & 940\\
people & 852\\
\midrule
\B{Total} & 25724\\
\bottomrule
\end{tabular}
\caption{NEWS categorization dataset label distribution.}
\label{table:rw_news}
\end{table}


\begin{table}[ht!]
\centering
\begin{tabular}{l l}
\toprule
\textbf{Score} & \textbf{Translation quality}\\
\midrule
1 & Invalid or meaningless translation \\
2 & Invalid but not totally wrong \\
3 & Almost valid, but not totally correct \\
4 & Valid and correct translation \\
\bottomrule
\end{tabular}
\caption{Machine-translated GLUE benchmark scoring prompt levels.}
\label{table:glue_scores}
\end{table}

\clearpage

\begin{table*}[!ht]
\centering
\resizebox{0.95\textwidth}{!}{%
\def\arraystretch{1.2}
\begin{tabular}{l r l l}
\toprule
\textbf{POS Tag} & \textbf{$\tilde{P}_p$ weight} & \textbf{Description} & \textbf{Example}  \rule{0pt}{2.8ex} \rule[-1.2ex]{0pt}{0pt}\\
\midrule
\textbf{V\#000} & 1.8 & Infinitive Verb & kuvuga \textit{`to say'}\\
\textbf{V\#001} & 1 & Gerund or verbal noun & uwavuze \textit{`the one who said'}\\
\textbf{V\#002} & 1.5 & Imperative verb & vuga \textit{`say'}\\
\textbf{V\#004} & 1.5 & Continuous present verb & aracyavuga \textit{`she is still saying'}\\
\textbf{V\#005} & 1.5 & Past tense verb & yaravuze \textit{`she said'}\\
\textbf{V\#006} & 1.5 & Future tense verb & azavuga \textit{`she will say'}\\
\textbf{V\#010} & 1.5 & Verb without tense mark & avuga \textit{`saying'}\\
\textbf{N\#011} & 1 & Noun without augmment & (wa)muntu \textit{`person'}\\
\textbf{N\#012} & 2 & Noun with augment & umuntu \textit{`a person'}\\
\textbf{DE\#013} & 2 & Demonstrative ng- & nguyu \textit{`this is her'}\\
\textbf{DE\#020} & 3 & Personal demonstrative & wowe \textit{`you'}\\
\textbf{DE\#021} & 2 & Demonstrative with augment & uwo \textit{`this (person)'}\\
\textbf{PO\#025} & 2 & Possessive +augment +owner & uwawe \textit{`yours'}\\
\textbf{QA\#026} & 0.5 & Qualificative adjective +augment +bu & ubuto \textit{`littleness'}\\
\textbf{QA\#027} & 1 & Qualificative adjective +augment -bu & umuto \textit{`the little one'}\\
\textbf{QA\#028} & 2.5 & Qualificative adjective -augment & muto \textit{`little'}\\
\textbf{QA\#029} & 3 & Qualificative adjective -augment +reduplication & mutomuto \textit{`(kind of) little'}\\
\textbf{NU\#030} & 2.5 & Numeral & babiri \textit{`two (people)'}\\
\textbf{OT\#033} & 2.5 & Quoting -ti & bati: \textit{`they said:'}\\
\textbf{NP\#035} & 2 & Proper names & Yohana \textit{`John'}\\
\textbf{DI\#036} & 3 & Digits & 84 \\
\textbf{AD\#037} & 2.5 & Adverb & bucece \textit{`silently'}\\
\textbf{VC\#038} & 2.5 & Conjunctive adverbs & hanyuma \textit{`and then'}\\
\textbf{CO\#039} & 2.5 & Commanding expressions & cyono \textit{`please'}\\
\textbf{CA\#040} & 2.5 & Calling expressions & yewe \textit{`you'}\\
\textbf{QU\#044} & 3 & Questioning adverb he & he \textit{`where'}\\
\textbf{SP\#054} & 2.5 & Spatial & hakurya \textit{`over there'}\\
\textbf{TE\#055} & 2.5 & Temporal & kare \textit{`early'}\\
\textbf{RL\#056} & 3 & Relatives & masenge \textit{`my aunt'}\\
\textbf{PR\#057} & 3 & Prepositions & ku \textit{`on'}\\
\textbf{OR\#064} & 2.5 & Orientations & amajyaruguru \textit{`north'}\\
\textbf{AJ\#065} & 2.5 & Adjectives & rusange \textit{`common'}\\
\textbf{NN\#066} & 2.5 & Nominal loanwords & kopi \textit{`copy'}\\
\textbf{HR\#067} & 3 & Hours & (saa) mbiri \textit{`eight o’clock'}\\
\textbf{DT\#068} & 2.5 & Date & taliki \textit{`date'}\\
\textbf{EN\#069} & 3 & Common English terms & live, like, share\\
\textbf{IJ\#070} & 2.5 & Interjections & dorere \textit{`see!'}\\
\textbf{CJ\#071} & 3 & Conjunctions & ko \textit{`that'}\\
\textbf{CP\#078} & 3 & Copula & ni \textit{`it is'}\\
\textbf{RE\#079} & 3 & Responses & yego \textit{`yes'}\\
\textbf{UN\#083} & 3 & Measuring units & metero \textit{`meter'}\\
\textbf{MO\#084} & 4 & Months & Mutarama \textit{`January'}\\
\textbf{PT\#085} & 3 & Punctuations & . \\
\bottomrule
\end{tabular}
}
\caption{Examples of POS tags used in KinyaBERT along with precedence weights $\tilde{P}_p(x_t|y_t)$ in {\eqnref{factored}}.}
\label{table:pos_tags_examples}
\end{table*}



\begin{table*}[!ht]
\centering
\resizebox{0.7\textwidth}{!}{%
\def\arraystretch{1.2}
\begin{tabular}{l l l}
\toprule
\textbf{Affix Set} & \textbf{Example}  & \textbf{Surface form}  \rule{0pt}{2.8ex} \rule[-1.2ex]{0pt}{0pt}\\
\midrule
\textbf{V:2:ku-V:18:a} & \textbf{ku-gend-a} & kugenda \textit{`to walk'}\\
\textbf{N:0:u-N:1:mu} & \textbf{u-mu-ntu} & umuntu \textit{`a person'}\\
\textbf{PO:1:i} & \textbf{i-a-cu} & yacu \textit{`our'}\\
\textbf{N:0:i-N:1:n} & \textbf{i-n-kiko} & inkiko \textit{`courts'}\\
\textbf{PO:1:u} & \textbf{u-a-bo} & wabo \textit{`their'}\\
\textbf{V:2:a-V:4:a-V:18:ye} & \textbf{a-a-bon-ye} & yabonye \textit{`she saw'}\\
\textbf{DE:1:u-DE:2:u} & \textbf{u-u-o} & uwo \textit{`that'}\\
\textbf{V:2:u-V:4:a-V:17:w-V:18:ye} & \textbf{u-a-vug-w-ye} & wavuzwe \textit{`who was talked about'}\\
\textbf{QA:1:ki-QA:3:ki-QA:4:re} & \textbf{ki-re-ki-re} & kirekire \textit{`tall'}\\
\bottomrule
\end{tabular}
}
\caption{Examples of affix sets used by KinyaBERT$_{ASC}$; there are 34K sets in total.}
\label{table:affix_set_examples}
\end{table*}


\begin{table*}[ht!]
\centering
\resizebox{\textwidth}{!}{%
\begin{tabular}{l c c c c c c c c}
\toprule
\B{Hyperparameter} & \B{MRPC} & \B{QNLI} & \B{RTE} & \textbf{SST-2}  & \textbf{STS-B}   & \textbf{WNLI}   & \textbf{NER}   & \textbf{NEWS} \\
\midrule
Peak Learning Rate & 1e-5 & 1e-5 & 2e-5 & 1e-5 & 2e-5 & 1e-5 & 5e-5 & 1e-5 \\
Batch Size & 16 & 32 & 16 & 32 & 16 & 16 & 32 & 32\\
Learning Rate Decay & Linear & Linear & Linear & Linear & Linear & Linear & Linear & Linear\\
Weight Decay & 0.1 & 0.1 & 0.1 & 0.1 & 0.1 & 0.1 & 0.1 & 0.1\\
Max Epochs & 15 & 15 & 15 & 15 & 15 & 15 & 30 & 15 \\
Warmup Steps proportion & 6\%  & 6\%  & 6\%  & 6\%  & 6\%  & 6\%  & 6\%  & 6\% \\
Optimizer & AdamW  & AdamW  & AdamW  & AdamW  & AdamW  & AdamW  & AdamW  & AdamW \\
\bottomrule
\end{tabular}
}
\caption{Downstream task fine-tuning hyper-parameters.}
\label{table:finetune_hyperpar}
\end{table*}

\begin{table*}[!ht]
\centering
\resizebox{\textwidth}{!}{%
\begin{tabular}{l l l l l}
\toprule
\multirow{2}{11em}{\textbf{Paper}} & \multirow{2}{3em}{\textbf{Language}} & \textbf{Pre-training} &  \textbf{Positional}&  \textbf{Input}\\
& & \textbf{Tasks} & \textbf{Embedding} & \textbf{Representation} \\
\midrule
\citet{mohseni-tebbifakhr-2019-morphobert} & Persian & MLM+NSP & Absolute & Morphemes\\

\citet{kuratov2019adaptation} & Russian & MLM+NSP & Absolute & BPE\\

\citet{masala2020robert} & Romanian & MLM+NSP & Absolute & BPE\\

\citet{baly2020arabert} & Arabic & WWM+NSP & Absolute & BPE\\

\citet{koto2020indolem} & Indonesian & MLM+NSP & Absolute & BPE\\

\citet{chan-etal-2020-germans} & German & WWM &  Absolute & BPE\\

\citet{delobelle-etal-2020-robbert} & Dutch & MLM & Absolute & BPE\\

\citet{nguyen-tuan-nguyen-2020-phobert} & Vietnamese & MLM &  Absolute & BPE\\

\citet{canete2020spanish} & Spanish & WWM & Absolute & BPE\\

\citet{rybak2020klej} & Polish & MLM & Absolute & BPE\\

\citet{martin-etal-2020-camembert}& French & MLM & Absolute & BPE\\

\citet{le2020flaubert} & French & MLM & Absolute & BPE\\

\citet{koutsikakis2020greek} & Greek & MLM+NSP & Absolute & BPE\\

\citet{souza2020bertimbau} & Portuguese & MLM & Absolute & BPE\\

\citet{ralethe2020adaptation} & Afrikaans & MLM+NSP & Absolute & BPE\\
\midrule
This work & Kinyarwanda & MLM: STEM+AFFIXES & TUPE-R & Morphemes+BPE\\
\bottomrule
\end{tabular}
}
\caption{Comparison between KinyaBERT and other monolingual BERT-variant PLMs. We only compare with previous works that have been published in either journals or conferences as of August 2021. We excluded some extremely high-resource languages such as English and Chinese. MLM: Masked language model; NSP: Next Sentence Prediction; WWM: Whole Word Masked.}
\label{table:monobert2}
\end{table*}

\clearpage

\begin{figure*}[!ht]
 \centering
\resizebox{\textwidth}{!}{%
 \begin{tabular}{c}
\resizebox{\textwidth}{!}{%
    \includegraphics{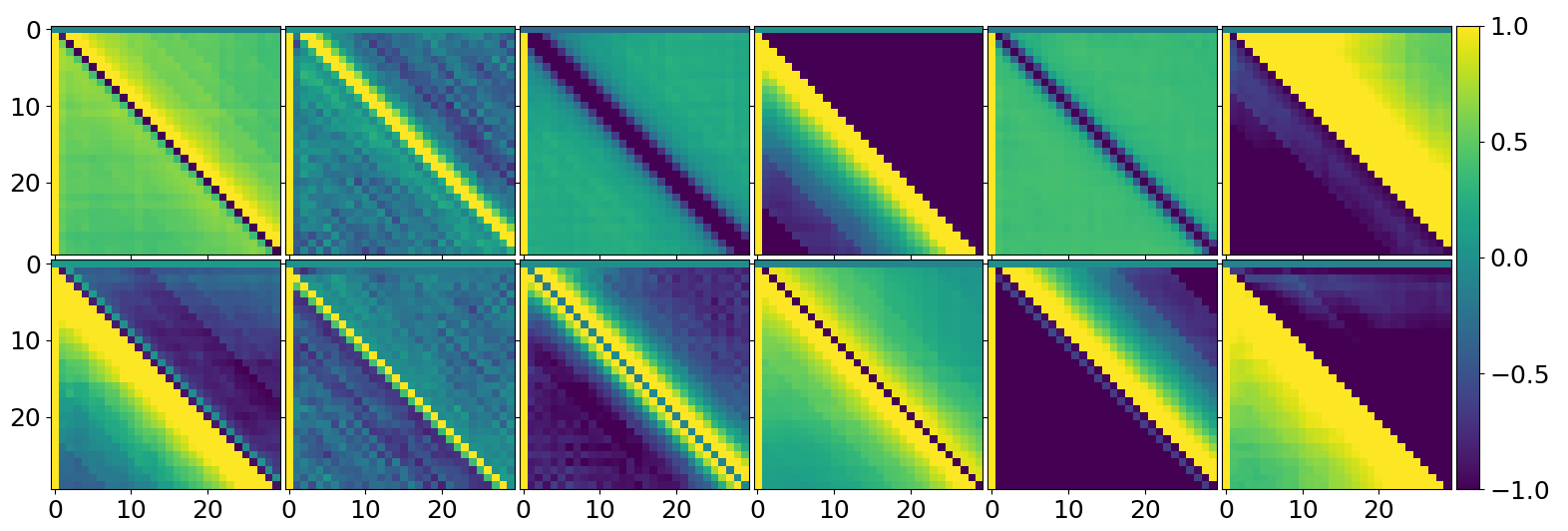}
    } \\
    BERT$_{BPE}$; Average non-adjacent diagonal STDEV = 0.81 for $|i-j|\in[2, 10]$ \\
\resizebox{\textwidth}{!}{%
    \includegraphics{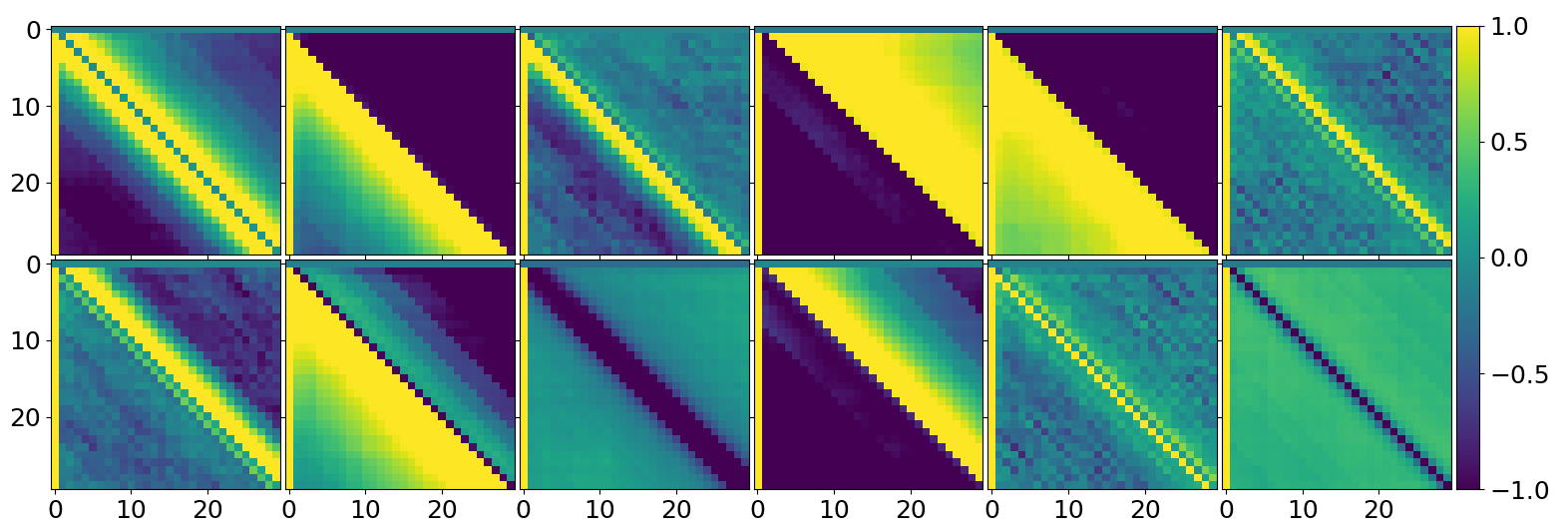}
    } \\
    BERT$_{MORPHO}$; Average non-adjacent diagonal STDEV = 0.80 for $|i-j|\in[2, 10]$ \\
\resizebox{\textwidth}{!}{%
    \includegraphics{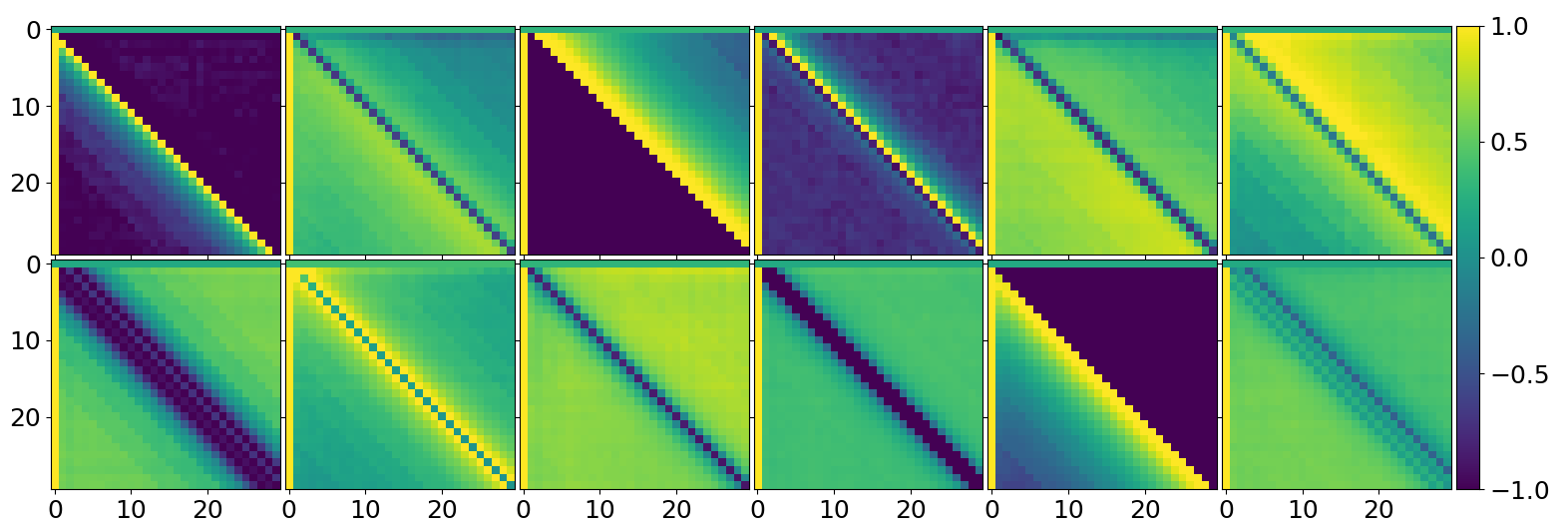}
    } \\
    KinyaBERT$_{ADR}$; Average non-adjacent diagonal STDEV = 0.75 for $|i-j|\in[2, 10]$ \\
\resizebox{\textwidth}{!}{%
    \includegraphics{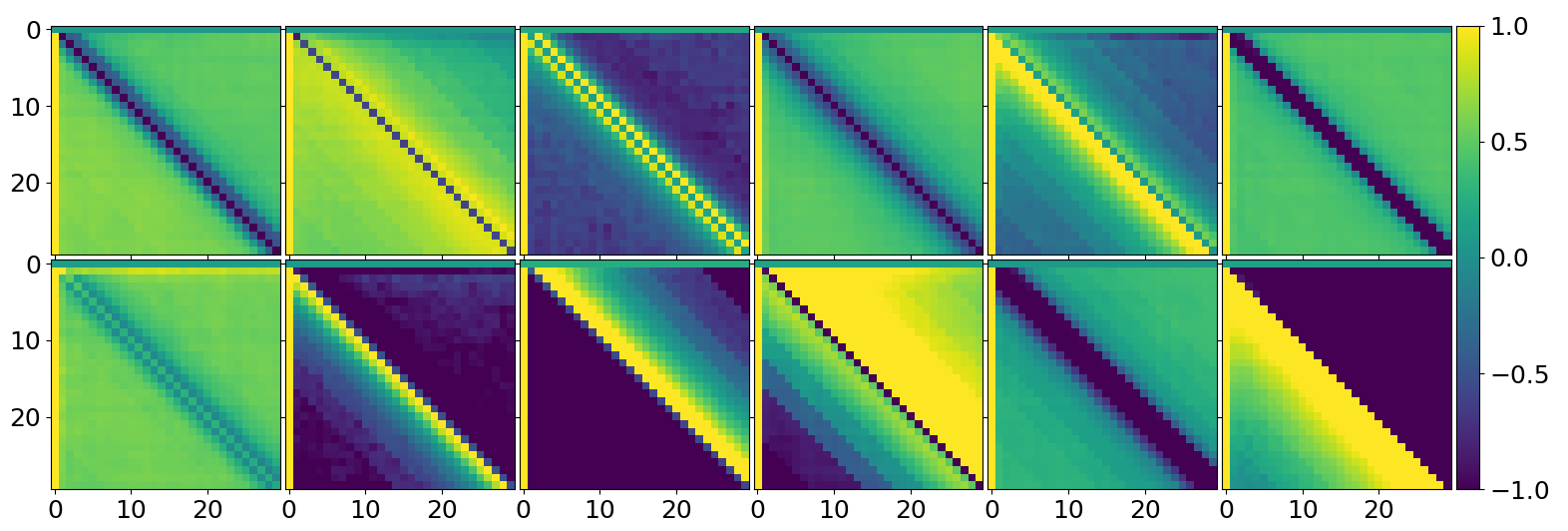}
    } \\
    KinyaBERT$_{ASC}$; Average non-adjacent diagonal STDEV = 0.75 for $|i-j|\in[2, 10]$\\
\end{tabular}
}
 \captionof{figure}{Visualization of the positional attention bias (normalized) of the 12 attention heads. Each $(i,j)$ attention bias~\cite{ke2020rethinking} indicates the positional correlations between the $i^{th}$ and $j^{th}$ words/tokens in a sentence.}
 \label{fig:pos_bias}
\end{figure*}

\end{appendices}

\end{document}